\documentclass[a4paper,twoside]{article}

\usepackage{epsfig}
\usepackage{subfigure}
\usepackage{calc}
\usepackage{amssymb}
\usepackage{amstext}
\usepackage{amsmath}
\usepackage{amsthm}
\usepackage{multicol}
\usepackage{pslatex}
\usepackage{apalike}
\usepackage{ulem}
\usepackage[usenames, dvipsnames]{color}
\usepackage[export]{adjustbox}
\usepackage{SCITEPRESS}     
\usepackage[ruled]{algorithm2e}
\begin{document}

\title{Semantic Image Inpainting Through Improved Wasserstein Generative Adversarial Networks}

\author{\authorname{Patricia Vitoria \textsuperscript{*}, Joan Sintes \thanks{These two authors contributed equally} and Coloma Ballester}
\affiliation{Department of Information and Communication Technologies, University Pompeu Fabra, Barcelona, Spain }
\email{\{patricia.vitoria, coloma.ballester\}@upf.edu, joansintesmarcos@gmail.com}
}

\abstract{Image inpainting is the task of filling-in missing regions of a damaged or incomplete image. In this work we tackle this problem not only by using the available visual data but also by incorporating image semantics through the use of generative models.
Our contribution is twofold: First, we learn a data latent space by training an improved version of the Wasserstein generative adversarial network, for which we incorporate a new generator and discriminator architecture. Second, the learned semantic information is combined with a new optimization loss for inpainting whose minimization infers the missing content conditioned by the available data. It takes into account powerful contextual and perceptual content inherent in the image itself.
The benefits include the ability to recover large regions by accumulating semantic information even it is not fully present in the damaged image. Experiments show that the presented method obtains qualitative and quantitative top-tier results in different experimental situations and also achieves accurate photo-realism comparable to state-of-the-art works.}

\onecolumn \maketitle \normalsize \vfill

\section{\uppercase{Introduction}}
\label{sec:introduction}
\noindent 
The goal of image inpainting methods is to recover missing information of occluded, missing or corrupted areas of an image in a realistic way, in the sense that the resulting image appears as of a real scene. Its applications are numerous and range from the automatization of cinema post-production tasks enabling, e.g., the deletion of annoying objects, to new view synthesis  generation for, e.g., broadcasting of sport events.

\begin{figure}
    \centering
    \begin{tabular}{cc}
   \includegraphics[width=3.0cm]{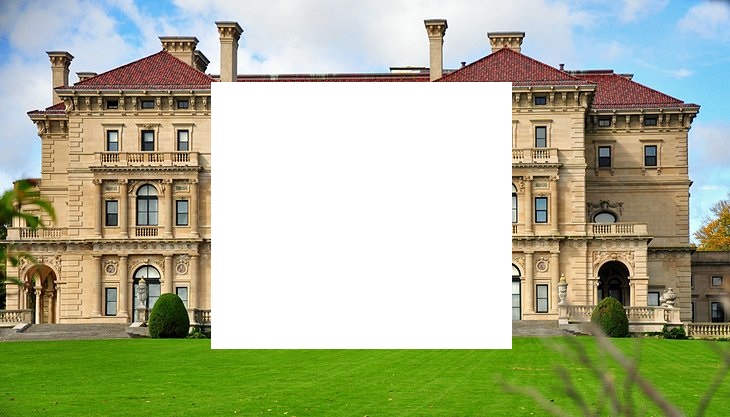}&
   \includegraphics[width=3.0cm]{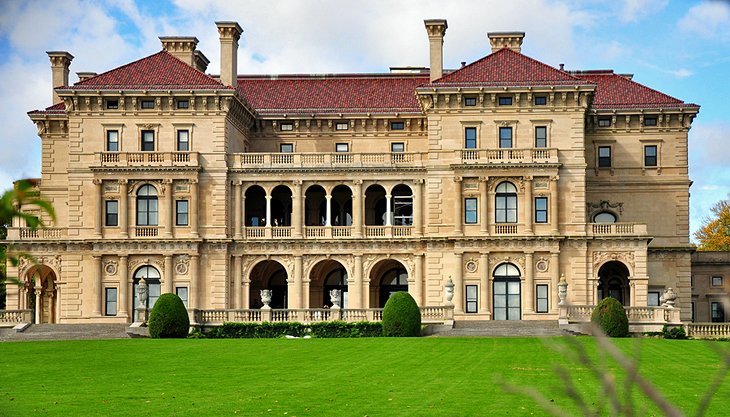} \\
    (a) & (b) \\
       \includegraphics{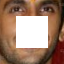} &
   \includegraphics{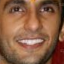} \\
    (c) & (d) 
    \end{tabular}
    \caption{Qualitative illustration of the task. Given the visible content in (a), our experience indicates that one or more central doors would be expected in such incomplete building. Thus, a plausible completion would be the one of (b). Also, our brain automatically {\it{completes}} the image in (c) with {\it{a face}} such as (d).}
    \label{fig:f1}
\end{figure}

Interestingly, it is a pervasive and easy task for a human to infer hidden areas of an image. Given an incomplete image, our brain unconsciously reconstructs the captured real scene by completing the gaps (called holes or inpainting masks in the inpainting literature). On the one hand, it is acknowledged that local geometric processes and global ones (such as the ones associated to geometry-oriented and exemplar-based models, respectively) are leveraged in the humans' completion phenomenon. On the other hand, humans use the experience and previous knowledge of the surrounding world to infer from memory what fits the context of a missing area. Figure \ref{fig:f1} displays two examples of it; looking at the image in Figure \ref{fig:f1}(a), our experience indicates that one or more central doors would be expected in such an incomplete building and, thus, a plausible completion would be the one of (b). Also, our {\it{trained}} brain automatically completes Figure \ref{fig:f1}(c) with the missing parts of {\it{a face}} such as the one shown in (d). 

Mostly due to its inherent ambiguity and to the complexity of natural images, the inpainting 
problem remains theoretically and computationally challenging, specially if large regions are missing. Classical methods use redundancy of the incomplete input image: smoothness priors in the case of geometry-oriented approaches and self-similarity principles in the non-local or exemplar-based ones.
Instead, using the terminology of~\cite{cont-enc,11}, semantic inpainting refers to the task of inferring arbitrary large missing regions in images based on image semantics. Applications such as the identification of different objects which were jointly occluded in the captured scene, 2D to 3D conversion, or image editing (in order to, e.g., removing or adding objects and changing the object category) could benefit from accurate semantic inpainting methods. Our work fits in this context. We capitalize on the understanding of more abstract and high level information that unsupervised learning strategies may provide.
%

Generative methods that produce novel samples from high-dimensional data distributions, such as
images, are finding widespread use, for instance in image-to-image translation \cite{im-to-im,liu2017unsupervised}, image synthesis and semantic manipulation \cite{wang2018high}, to mention but a few. Currently the most prominent approaches include autoregressive models \cite{autoagressive}, variational autoencoders (VAE) \cite{vae}, and generative adversarial networks \cite{GAN}. Generative Adversarial Networks (GANs) are often credited for producing less burry outputs when used for image generation. It consists of a framework for training generative parametric models based on a game between two networks: a generator network that produces synthetic data from a noise source and a discriminator network that differentiates between the output of the genererator and true data. The approach has been shown to produce high quality images and even videos \cite{zhu2017unpaired,pumarola2018unsupervised,chan2018everybody}. 

We present a new method for semantic image inpainting with an improved version of the Wasserstein GAN \cite{WGAN} 
including a new generator and discriminator architectures and  a novel optimization loss in the context of semantic inpainting that outperforms related approaches. 
More precisely, our contributions are summarized as follows:
\begin{itemize}
    \item We propose several improvements to the architecture based on an improved WGAN such as the introduction of the residual learning framework in both the generator and discriminator, the removal of the fully connected layers on top of convolutional features and the replacement of the widely used batch normalization by a layer normalization. 
    These improvements ease the training of the networks making them to be deeper and stable.
    \item We define a new optimization loss 
    that takes into account, on the one side, the semantic information inherent in the image, and, on the other side, contextual information that capitalizes on the image values and gradients.
    \item We quantitatively and qualitatively show that our proposal achieves top-tier results on two datasets: CelebA and Street View House Numbers.
\end{itemize}
The remainder of the paper is organized as follows. In Section \ref{sec:sota}, we review the related state-of-the-art work focusing first on generative adversarial networks and then on inpainting methods. Section  \ref{sec:method} details our whole method. In Section \ref{sec:experimental_results}, we present both quantitative and qualitative assessments of all parts of the proposed method. Section \ref{sec:conclusions} concludes the paper.

\begin{figure}
    \centering
    \begin{tabular}{c}
      \includegraphics[width=5cm]{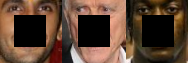} \\
    (a) \\    \includegraphics[width=5cm]{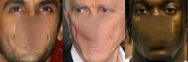} \\
       (b) \\
      \includegraphics[width=5cm]{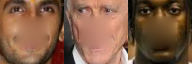} \\
         (c) \\
      \includegraphics[width=5cm]{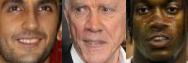} \\
      (d)
    \end{tabular}
\caption{Image inpainting results using three different approaches. (a) Input images, each with a big hole or mask. (b) Results obtained with the non-local method \cite{non-local}. (c) Results with the local method \cite{local_ipol}. (d) Our semantic inpainting method.}
\label{fig:f2}
\end{figure}

\section{\uppercase{Related Work}}
\label{sec:sota}
\paragraph{Generative Adversarial Networks.} 
GAN learning strategy \cite{GAN} is based on a game theory scenario between two networks, the generator's network and the discriminator's network, having adversarial objectives. The generator maps a source of noise from the latent space to the input space and the discriminator receives either a generated or a real image and must distinguish between both. The goal of this training procedure is to learn the parameters of the generator so that its probability distribution is as closer as possible to the one of the real data. To do so, the discriminator $D$ is trained to maximize the probability of assigning the correct label 
to both real examples and samples from the generator $G$, while $G$ is trained to fool the discriminator and to minimize $\log(1-D(G(z)))$ by generating realistic examples.
In other words, $D$ and $G$ play the following min-max game with value function $V(G,D)$ defined as follows:
\begin{equation}\label{eq:1}
\begin{split}
    \min_{G}\max_DV(D,G) = &\mathop{\mathbb{E}}_{x\sim P_{data}(x)}[\log D(x)] \\ & + \mathop{\mathbb{E}}_{z\sim p_{z}(z)}[\log (1-D(G(z)))]
\end{split}
\end{equation}
The authors of \cite{DCGAN} introduced convolutional layers to the GANs architecture, and proposed the so-called Deep Convolutional Generative Adversarial Network (DCGAN). GANs have been applied with success to many specific tasks such as image colorization \cite{9}, text to image synthesis \cite{text-to-image}, super-resolution \cite{14}, image inpainting \cite{11,13,12}, and image generation \cite{DCGAN,LSGAN,WGAN-GP,nguyen2016plug}, to name a few.
However, three difficulties still persist as challenges. One of them is the quality of the generated images 
and the remaining two are related to the well-known instability problem in the training procedure. Indeed,
two problems can appear: vanishing gradients and mode collapse.
Vanishing gradients are specially problematic  when comparing probability distributions with non-overlapping
supports. If the discriminator is able to perfectly distinguish between real and generated images, it reaches its optimum and thus the generator no longer improves the generated data. On the other hand, mode collapse happens when the generator only encapsulates the major nodes of the real distribution, and not the entire distribution. As a consequence, the generator keeps producing similar outputs to fool the discriminator.

Aiming a stable training of GANs, several authors have promoted the use of the  Wasserstein GAN (WGAN). WGAN minimizes
an approximation of the Earth-Mover (EM) distance or Wasserstein-1 metric between two probability distributions. The EM distance intuitively provides a measure of how much mass needs to be transported to transform one distribution into the other distribution. The authors of \cite{WGAN} analyzed the properties of this distance. They showed that one of the main benefits of the Wasserstein distance is that it is continuous. This property allows to robustly learn a probability distribution by smoothly modifying the parameters through gradient descend. Moreover, the Wasserstein or  EM distance is known to be a powerful tool to
compare probability distributions with non-overlapping
supports, in contrast to other distances such as the Kullback-Leibler divergence and the Jensen-Shannon divergence (used in the DCGAN and other GAN approaches) which produce the vanishing gradients problem, as mentioned above. Using the Kantorovich-Rubinstein duality, the Wasserstein distance between two distributions, say a {\it{real}} distribution $P_{real}$ and an estimated distribution $P_g$, can be computed as
\begin{equation}\label{eq2}
W(P_{real},P_g) = \text{sup}\,  {\mathbb{E}}_{x\sim P_{real}}\left[ f(x)\right] - {\mathbb{E}}_{x\sim P_{g}}\left[ f(x)\right]
\end{equation}
where the supremum is taken over all the 1-Lipschitz functions $f$ (notice that, if $f$ is differentiable, it implies that ${ \|\nabla f\|\leq 1}$). 
Let us notice that $f$ in Equation \eqref{eq2} can be thought to take the role of the discriminator $D$ in the GAN terminology. 
In \cite{WGAN}, the Wasserstein GAN is defined as the network whose parameters are learned through optimization of 
\begin{equation}\label{eq3}
\min_G \max_{D\in{\cal D}} \, {\mathbb{E}}_{x\sim P_{real}}\left[ D(x)\right] - {\mathbb{E}}_{x\sim P_{G}}\left[ D(x)\right] 
\end{equation}
where ${\cal{D}}$ denotes the set of 1-Lipschitz functions. Under an optimal discriminator (called a \textit{critic} in \cite{WGAN}), minimizing the value function with respect to the generator parameters minimizes $W(P_{real},P_g)$. To enforce the Lipschitz constraint, the authors proposed to use an appropriate weight clipping.  
The resulting WGAN solves the vanishing problem, but several authors \cite{WGAN-GP,BWGAN} have noticed that weight clipping is not the best solution to enforce the Lipschitz constraint and it causes optimization difficulties. For instance, the WGAN discriminator ends up learning an extremely simple function and not the real distribution. Also, the clipping threshold must be properly adjusted. Since a differentiable function is 1-Lipschitz if it has gradient with norm at most 1 everywhere, \cite{WGAN-GP} proposed an alternative to weight clipping: To add a gradient penalty term constraining the $L^2$ norm of the gradient while optimizing the original WGAN during training. 
Recently, the Banach Wasserstein GAN (BWGAN) \cite{BWGAN} has been proposed extending WGAN implemented via a gradient penalty term to any separable complete normed space. In this work we leverage the mentioned WGAN \cite{WGAN-GP} 
improved with a new design of the generator and discriminator architectures.

\paragraph{Image Inpainting.}
Most inpainting methods found in the literature can be classified into two groups: model-based approaches and deep learning approaches. In the former, two main groups can be distinguished: local and non-local methods. In local methods, also denoted as geometry-oriented methods, images are modeled as functions with some degree of smoothness. 
\cite{masnoumorel1998,inpaintingTV_chan_2001,BBCSV-01,local_ipol,
inpaintingCao2011}. These methods show good performance in propagating smooth level lines or gradients,  but fail in the presence of texture or for large missing regions. Non-local methods (also called exemplar- or patch-based) 
exploit the self-similarity prior by directly sampling the desired texture to perform the synthesis 
\cite{texture,inpainting_demanet_2003,inpainting_criminisi_2004,wang_affine_inpainting_2008,inpainting_kawai_2009,inpainting_aujol_2010,inpainting_arias_2011,huang_tog_2014,fedorov2016affine}. 
They provide impressive results in inpainting textures and repetitive structures even in the case of large holes. However, both type of methods use redundancy of the incomplete input image: smoothness priors in the case of geometry-based and self-similarity principles in the non-local or patch-based ones. Figures \ref{fig:f2}(b) and (c) illustrate the inpainting results (the inpaining hole is shown in (a)) using a local method (in particular \cite{local_ipol}) and the non-local method \cite{non-local}, respectively. As expected, the use of image semantics improve the results, as shown in (d).

Current state-of-the-art is based on deep learning approaches 
\cite{11,12,cont-enc,high-res,yu2018generative}.  \cite{cont-enc} modifies the original GAN architecture by inputting the image context instead of random noise to predict the missing patch. They proposed an encoder-decoder network using the combination of the $L^2$ loss and the adversarial loss and applied adversarial training to learn features while regressing the missing part of the image. \cite{11} proposes a method for semantic image inpainting, which generates the missing content by conditioning on the available data given a trained generative model.  In \cite{high-res}, a method is proposed to tackle inpainting of large parts on large images. 
They adapt multi-scale techniques to generate high-frequency details on top of the reconstructed object to achieve high resolution results.  Two recent works 
\cite{face-gen,glob-loc} add a discriminator network that considers only the filled region to emphasize the adversarial loss on top of the global GAN discriminator (G-GAN). This additional network, which is called the local discriminator (L-GAN), facilitates exposing the local structural details. Also, \cite{12} designs a discriminator that aggregates the local and global information by combining a G-GAN and a Patch-GAN that first shares network layers and later uses split paths with two separate adversarial losses in order to capture both local continuity and holistic features in the inpainted images. 
\begin{figure*}[ht]
   \centering
   \includegraphics[width=15.5cm]{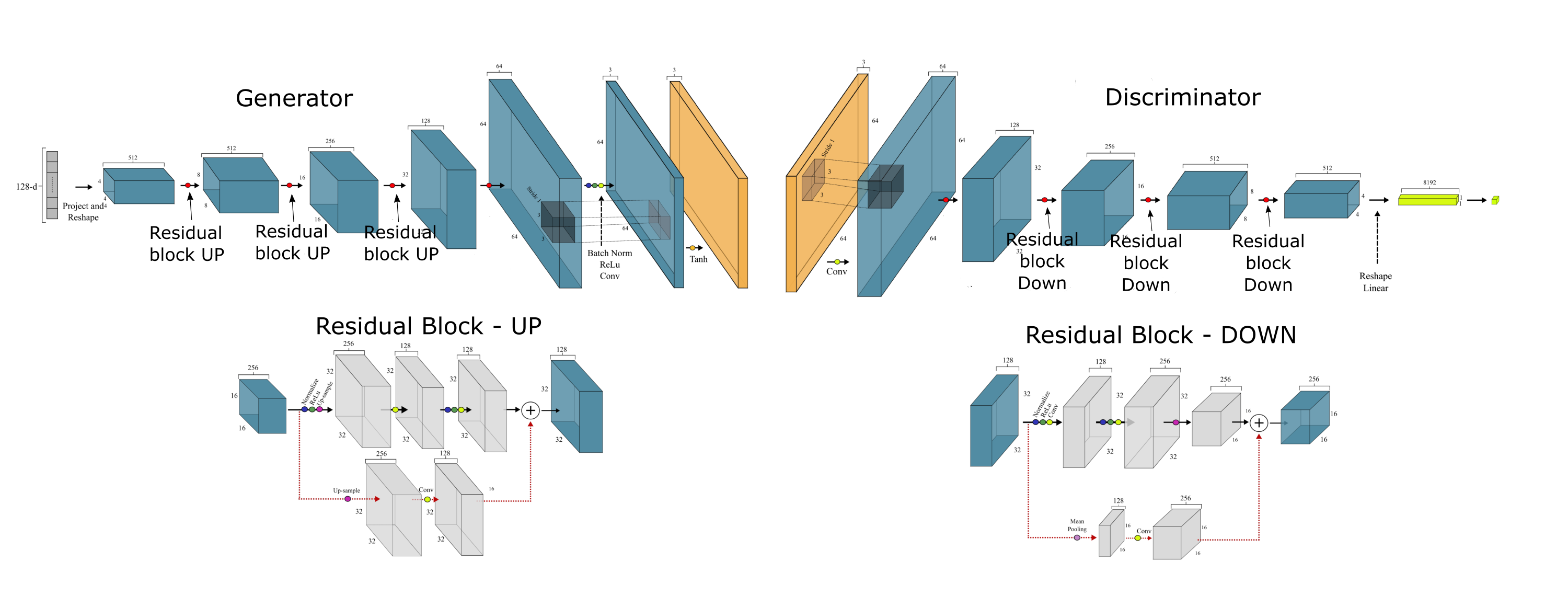}
   \caption{Overview  of the architecture of our improved WGAN. Top: generator and discriminator architectures (left and right, respectively). Bottom: corresponding residual block strategies}
   \label{fig:architecture} 
\end{figure*}

\section{\uppercase{Proposed Method}}
\label{sec:method}
\noindent Our semantic inpainting method is built on two main blocks: First, given a dataset of (non-corrupted) images, we train an improved version of the Wasserstein GAN to implicitly learn a data latent space to subsequently generate new samples from it. Then, given an incomplete image and the previously trained generative model,  we perform an iterative minimization procedure to infer the missing content of the incomplete image by conditioning on the known parts of the image. This procedure consists of the search of the closed encoding of the corrupted data in the latent manifold by minimization of a new loss which is made of a combination of contextual, through image values and image gradients, and prior losses. 

\subsection{Improved Wasserstein Generative Adversarial Network}
\label{subsec:architecture}
Our improved WGAN is built on the WGAN by \cite{WGAN-GP}, on top of which we propose several improvements. As mentioned above, the big counterpart of the generative models is their training instability which is very sensible not only to the architecture but also to the training procedure. In order to improve the stability of the network we propose several changes in its architecture. In the following we explain them in detail: 
\begin{itemize}
    \item First, network depth is of crucial importance in neural network architectures; using deeper networks more complex, non-linear functions can be learned, but deeper networks are more difficult to train.  In contrast to the usual model architectures of GANs, we have introduced in both the generator and discriminator the residual learning framework which eases the training of these networks, and enables them to be substantially deeper and stable. The degradation problem occurs when as the network depth increases, the accuracy  saturates (which might be unsurprising) 
    and then degrades rapidly. Unexpectedly, such degradation is not caused by overfitting, and adding more layers to a suitably deep model leads to higher training errors \cite{residual}. For that reason we have introduced residual blocks in our model. Instead of hoping each sequence of layers to directly fit a desired mapping, we explicitly let these layers fit a residual mapping. Therefore, the input $x$ of the residual block is recast into $F(x)+x$ at the output. At the bottom of Figure \ref{fig:architecture}, the layers that make up a residual block in our model are displayed. 
    \item Second, to eliminate fully connected layers on top of convolutional features is a widely used approach. Instead of using fully connected layers we directly connect the highest convolutional features to the input and the output, respectively, of the generator and discriminator. The first layer of our GAN generator, which takes as input a sample $z$ of a normalized Gaussian distribution, could be called fully connected as it is just a matrix multiplication, but the result is reshaped into a four by four 512-dimensional tensor and used as the start of the convolution stack. In the case of the discriminator, the last convolution layer is flattened into  a single scalar. Figure \ref{fig:architecture} displays a visualization of the  architecture of the generator (top left) and of the discriminator (top right). 
    \item Third, most previous GAN implementations use batch normalization in both the generator and the discriminator to help stabilize training. However, batch normalization changes the form of the discriminator's problem from mapping a single input to a single output to mapping from an entire batch of inputs to a batch of outputs \cite{train-gans}. Since we penalize the norm of the gradient of the critic (or discriminator) with respect to each input independently, and not the entire batch, we omit batch normalization in the critic. To not introduce correlation between samples, we use layer normalization \cite{layer-norm} as a drop-in replacement for batch normalization in the critic.
    \item Finally, the ReLU activation is used in the generator with the exception of the output layer which uses the Tanh function.  Within the discriminator we also use ReLu activation. This is in contrast to the DCGAN, 
    which makes use of the LeakyReLu.
\end{itemize}

\subsection{Semantic Image Completion}
\label{subsec:image_completion}
Once we have trained our generative model until the data latent space has been properly estimated from uncorrupted data, we perform semantic image completion. 
After training the generator $G$ and the discriminator (or critic) $D$, $G$ is able to take a random vector $z$ drawn from $p_z$ and generate an image mimicking samples from $P_{real}$. The intuitive idea is that if $G$ is efficient in its representation, then, an image that does not come from $P_{real}$, such as a corrupted image, should not lie on the learned encoding manifold of $z$. Therefore, our aim is to recover the encoding $\hat{z}$ that is closest to the corrupted image while being constrained to the manifold. Then, when $\hat{z}$ is found, we can restore the damaged areas of the image by using our trained generative model $G$ on $\hat{z}$.

\begin{figure}
    \centering
    \begin{tabular}{c}
       \includegraphics{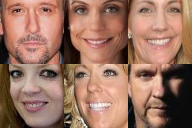} \\
    (a) \\    \includegraphics{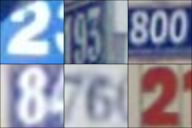} \\
       (b) \\
        \includegraphics{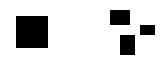} \\
       (c) \\
    \end{tabular}
\caption{(a) Samples from CelebA training dataset. (b) Samples from SVHN training dataset. (c) Two masks $M$ used in our experiments corresponding to different corrupted regions (the black pixels): A central square on the left and three rectangular areas on the right.}
\label{fig:datasetMask}
\end{figure}

We formulate the process of finding $\hat{z}$ as an optimization problem.  Let $y$ be a damaged image and $M$ a  binary mask of the same spatial size as the image, where the white pixels ($M(i)=1$) determine the uncorrupted areas of $y$. Figure \ref{fig:datasetMask}(c) shows two different masks $M$ corresponding to different corrupted regions (the black pixels): A central square on the left and three rectangular areas on the right. We define the closest encoding $\hat{z}$ as the optimum of following optimization problem with the new loss:
\begin{equation}
\label{eq:completion_loss}
\hat{z} = arg \mathop{min}_z\{\alpha  \mathcal{L}_c(z|y,M) + \beta  \mathcal{L}_g(z|y,M) + \eta \mathcal{L}_p(z)\}
\end{equation}
where $\alpha,\beta,\eta >0$, $\mathcal{L}_c$  and $\mathcal{L}_g$ are contextual losses constraining the generated image by the input corrupted image $y$ on the regions with available data given by $M$, 
and  $\mathcal{L}_p$ denotes the prior loss. In particular, the contextual loss $\mathcal{L}_c$ constrains the image values and the gradient loss $\mathcal{L}_g$ is designed to constraint the image gradients. 
More precisely, the contextual loss $\mathcal{L}_c$ is defined as the $L^1$ norm between the generated samples $G(z)$ and the uncorrupted parts of the input image $y$ weighted 
in such a way that the optimization loss pays more attention to the pixels that are close to the corrupted area when searching for the optimum encoding $\hat{z}$.
To do so, for each uncorrupted pixel $i$ in the image domain, we define its weight $W(i)$ as
\begin{equation}
\label{eq:weights}
W(i) =
\begin{cases}
\displaystyle\sum_{j \in N_i}  \frac{(1-M(j))}{|N_i|} & if   M(i) \neq 0\\
0&  if  M(i) = 0
\end{cases}
\end{equation}
where $N_i$ denotes a local neighborhood or window centered at $i$, and $|N_i|$ denotes its cardinality, i.e., the area (or number of pixels) of $N_i$. This weighting term was also used by \cite{11}. In order to provide a comparison with them, we use the same window size of 7x7 in all the experiments. Finally, we define the contextual loss $\mathcal{L}_c$ as
\begin{equation}
\label{eq:contextual_loss}
\mathcal{L}_c(z|y,M) = W  \|M  (G(z) - y)\|
\end{equation}
Our gradient loss $\mathcal{L}_g$ represents also a contextual term and it is defined as 
the $L^1$-norm of the difference between the gradient of the uncorrupted portion and the gradient of the recovered image, that is,
\begin{equation}
\label{eq:gradient_loss}
\mathcal{L}_g(z|y,M) = W \|M ( \nabla G(z) - \nabla y)\|
\end{equation}
where $\nabla$ denotes the gradient operator. The idea behind the proposed gradient loss is to constrain the structure of the generated image given the structure of the input corrupted image. The benefits are specially noticeable for a sharp and detailed inpainting of large missing regions which typically contain some kind of structure (e.g. nose, mouth, eyes, texture, etc, in the case of faces). In contrast, the contextual loss $\mathcal{L}_c$ gives the same importance to the homogeneous zones and structured zones and it is in the latter where the differences are more important and easily appreciated. In practice, the image gradient computation is approximated by central finite differences. In the boundary of the inpainting hole, we use either forward or backward differences depending on whether the non-corrupted information is available. 

Finally, the prior loss $ \mathcal{L}_p$ 
is defined such as it favours realistic images, similar to the samples that are used to train our generative model, that is,
\begin{equation}
\label{eq:prior_loss}
\mathcal{L}_p(z) = -D_w(G_{\theta}(z))
\end{equation}
where $D_w$ is the output of the discriminator $D$ with parameters $w$  given the image $G_{\theta}(z)$ generated by the generator $G$ with parameters $\theta$ and input vector $z$. In other words, the prior loss is defined as our second WGAN loss term in \eqref{eq3} penalizing unrealistic images.
Without $\mathcal{L}_p$ the mapping from $y$ to $z$ may converge to a perceptually implausible result. Therefore $z$ is updated to fool the discriminator and make the corresponding generated image more realistic.

\begin{figure}[ht]
\centering
   \includegraphics[width=7.5cm]{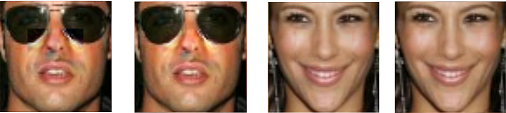}\\
   (a) $\qquad\qquad$ (b) $\qquad\qquad$  (c) $\qquad\quad$ (d) 
\caption{(b) and (d) show the results obtained after applying Poisson editing (equation \eqref{eq:poisson_editing} in the text) to the inpainting results shown in (a) and (c), respectively.}
\label{fig:f7} 
\end{figure}

The parameters  $\alpha$, $\beta$ and $\eta$  in equation (\ref{eq:completion_loss}) allow to balance among the three losses. The selected parameters are $\alpha = 0.1$,   $\beta = 1-\alpha$ and $\eta = 0.5$ but for the sake of a thorough analysis we present in Tables \ref{table:central_bloc} and \ref{table:treeSquares} an ablation study of our contributions. With the defined contextual, gradient and prior losses, the corrupted image can be mapped to the closest $z$ in the latent representation space, denoted by $\hat{z}$. $z$ is randomly initialized with Gaussian noise of zero mean and unit standard deviation and updated using back-propagation on the total loss given in the equation (\ref{eq:completion_loss}). 
Once $G(\hat{z})$ is generated, the inpainting result can be obtained by overlaying the uncorrupted pixels of the original damaged image to the generated image. Even so, the reconstructed pixels may not exactly preserve the same intensities of the surrounding pixels although the  content and structure is correctly well aligned. To solve this problem, a Poisson editing step \cite{poison} is added at the end of the pipeline in order to reserve the gradients of $G(\hat{z})$ without mismatching intensities of the input image $y$.  
Thus, the final reconstructed image $\hat{x}$ is equal to:
\begin{equation}
\label{eq:poisson_editing}
\begin{split}
\hat{x} =  arg \mathop{min}_x \|\nabla x - \nabla G(\hat{z})\|_2^2 \\
\text{ such that } x(i) = y(i) \text{ if } M(i) = 1
\end{split}
\end{equation}
Figure \ref{fig:f7} shows an example where visible seams are appreciated in (a) and (c), but less in (b) and (d) after applying \eqref{eq:poisson_editing}.

\section{\uppercase{Experimental Results}}
\label{sec:experimental_results}
\noindent
In this section we evaluate the proposed method both qualitatively and quantitatively by using different evaluation metrics. We compare our results with the results obtained by \cite{11} as both algorithms use first a GAN procedure to learn semantic information from a dataset and, second, combine it with an optimization loss for inpainting in order to infer the missing content. In order to perform an ablation study of all our contributions, we present the results obtained not only by using the original algorithm by \cite{11} but also the results obtained by adding our new gradient-based term $\mathcal{L}_g(z|y,M)$ to their original inpainting loss, and varying the trade-off between the different loss terms (weights $\alpha,\beta,\eta$). 

In the training step of our algorithm, we use the proposed architecture (see Section \ref{subsec:architecture}) where the generative model takes a random vector, of dimension 128, drawn from a normal distribution.  In contrast, \cite{11} uses the DCGAN architecture where the generative model takes a random 100 dimensional vector following a uniform distribution between $[-1, 1]$. For all the experiments we use: A fixed number of iterations equal to 50000, batch size 
equal to 64, learning rate 
equal to 0.0001 and exponential decay rate for the first and second moment estimates in the Adam update technique, $\beta_1=0,0$ and $\beta_2=0,9$, respectively. To increase the amount of training data we also performed data augmentation by randomly applying a horizontal flipping on the training set. Training the generative model required three days using an NVIDIA  TITAN X GPU.

In the inpainting step, the window size used to compute $W(i)$ in \eqref{eq:weights} is fixed to 7x7 pixels.
In our algorithm, we use back-propagation to compute $\hat{z}$ in the latent space. We make use of an Adam optimizer and restrict $z$ to $[-1, 1]$ in each iteration, which we found it produces more stable results. In that stage we used the Adam hyperparameters learning rate, $\alpha$, equal to 0.03 and the exponential decay rate for the first and second moment estimates, $\beta_1=0,9$ and $\beta_2=0,999$, respectively. After initializing with a random 128 dimensional vector $z$ drawn from a normal distribution, we perform 1000 iterations.

\begin{figure*}[h!]
\centering
   \includegraphics[width=15cm]{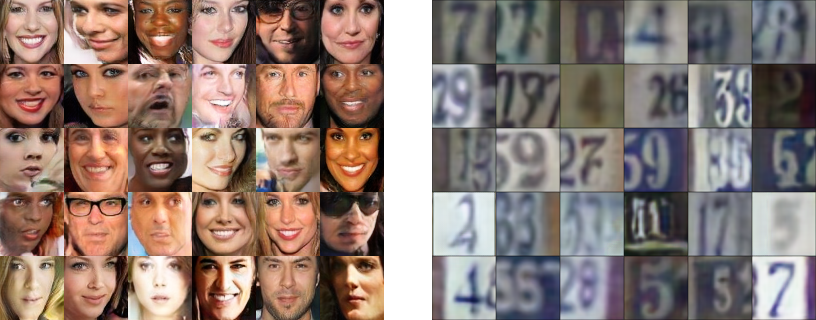}
\centering
\caption{Some images generated by our generative model using the CelebA and the SVHN dataset as training set, respectively. The CelebA dataset contains around 200k training images which are aligned and preprocessed to reduce the diversity between samples.  The SVHN dataset contains 73.257 training images. In this case, no pre-processing to reduce the diversity between samples has been applied. Notice that both datasets have been down-sampled to 64x64 pixel size before training.}
\label{fig:generatedDatasets} 
\end{figure*}

The assessment is given on two different datasets in order to check the robustness of our method: the CelebFaces Attributes Datases \cite{CelebA} and the Street View House Numbers (SVHN) \cite{SVHN}. CelebA dataset contains a total of 202.599 celebrity images covering large pose variations and background clutter. We split them into two groups: 201,599 for training and 1,000 for testing. In contrast, SVHN contains only 73,257 training images and 26,032 testing images. SVHN images are not aligned and have different shapes, sizes and backgrounds. The images of both datasets have been cropped with the provided bounding boxes and resized to only 64x64 pixel size. Figure \ref{fig:datasetMask}(a)-(b) displays some samples from these datasets. 

Let us remark that we have trained the proposed improved WGAN by using directly the images from the datasets without any mask application. Afterwards, our semanting inpainting method is evaluated on both datasets using the inpainting masks illustrated in Figure \ref{fig:datasetMask}(c). Notice that our algorithm can be applied to any type of inpainting mask.

\begin{figure*}[h]
    \centering
    \begin{tabular}{ccccccc}
                              \includegraphics[width=1.5cm]{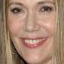} &
          \includegraphics[width=1.5cm]{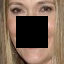} &
          \includegraphics[width=1.5cm]{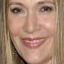} &
          \includegraphics[width=1.5cm]{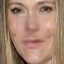} &
          \includegraphics[width=1.5cm]{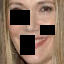} &
          \includegraphics[width=1.5cm]{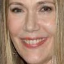} &
          \includegraphics[width=1.5cm]{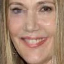} \\
                                        \includegraphics[width=1.5cm]{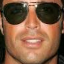} &
          \includegraphics[width=1.5cm]{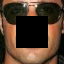} &
          \includegraphics[width=1.5cm]{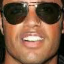} &
          \includegraphics[width=1.5cm]{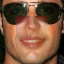} &
          \includegraphics[width=1.5cm]{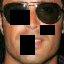} &
          \includegraphics[width=1.5cm]{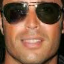} &
          \includegraphics[width=1.5cm]{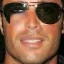} \\            \includegraphics[width=1.5cm]{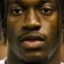} &
          \includegraphics[width=1.5cm]{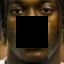} &
          \includegraphics[width=1.5cm]{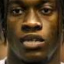} &
          \includegraphics[width=1.5cm]{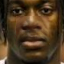} &
          \includegraphics[width=1.5cm]{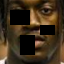} &
          \includegraphics[width=1.5cm]{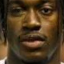} &
          \includegraphics[width=1.5cm]{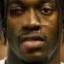} \\
          \includegraphics[width=1.5cm]{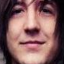} &
          \includegraphics[width=1.5cm]{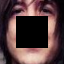} &
          \includegraphics[width=1.5cm]{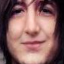} &
          \includegraphics[width=1.5cm]{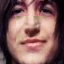} &
          \includegraphics[width=1.5cm]{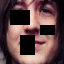} &
          \includegraphics[width=1.5cm]{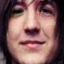} &
          \includegraphics[width=1.5cm]{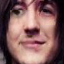} \\
                    \includegraphics[width=1.5cm]{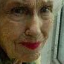} &
          \includegraphics[width=1.5cm]{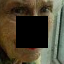} &
          \includegraphics[width=1.5cm]{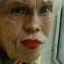} &
          \includegraphics[width=1.5cm]{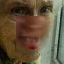} &
          \includegraphics[width=1.5cm]{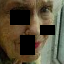} &
          \includegraphics[width=1.5cm]{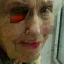} &
          \includegraphics[width=1.5cm]{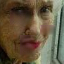} \\
          \includegraphics[width=1.5cm]{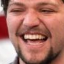} &
          \includegraphics[width=1.5cm]{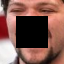} &
          \includegraphics[width=1.5cm]{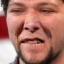} &
          \includegraphics[width=1.5cm]{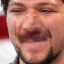} &
          \includegraphics[width=1.5cm]{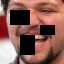} &
          \includegraphics[width=1.5cm]{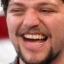} &
          \includegraphics[width=1.5cm]{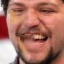} \\
          \includegraphics[width=1.5cm]{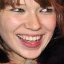} &
          \includegraphics[width=1.5cm]{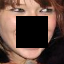} &
          \includegraphics[width=1.5cm]{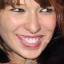} &
          \includegraphics[width=1.5cm]{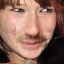} &
          \includegraphics[width=1.5cm]{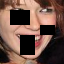} &
          \includegraphics[width=1.5cm]{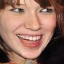} &
          \includegraphics[width=1.5cm]{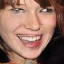} \\

    Original & Masked & Ours & SIMDGM & Masked & Ours & SIMDGM
    \end{tabular}
\caption{Inpainting results on the CelebA dataset: Qualitative comparison with the method \cite{11} (fourth and seventh columns, referenced as SIMDGM), using the two masks shown in the second and fifth columns, is also displayed.}
\label{fig:FinalResultsCelebA}
\end{figure*}

\begin{figure*}
    \centering
    \begin{tabular}{ccccccc}
          \includegraphics[width=1.5cm]{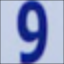} &
          \includegraphics[width=1.5cm]{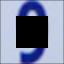} &
          \includegraphics[width=1.5cm]{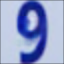} &
          \includegraphics[width=1.5cm]{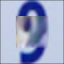} &
          \includegraphics[width=1.5cm]{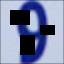} &
          \includegraphics[width=1.5cm]{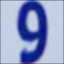} &
          \includegraphics[width=1.5cm]{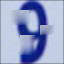} \\
            \includegraphics[width=1.5cm]{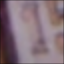} &
          \includegraphics[width=1.5cm]{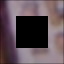} &
          \includegraphics[width=1.5cm]{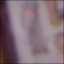} &
          \includegraphics[width=1.5cm]{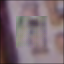} &
          \includegraphics[width=1.5cm]{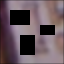} &
          \includegraphics[width=1.5cm]{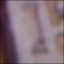} &
          \includegraphics[width=1.5cm]{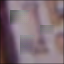} \\
         \includegraphics[width=1.5cm]{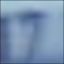} &
          \includegraphics[width=1.5cm]{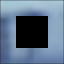} &
          \includegraphics[width=1.5cm]{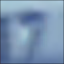} &
          \includegraphics[width=1.5cm]{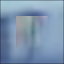} &
          \includegraphics[width=1.5cm]{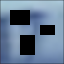} &
          \includegraphics[width=1.5cm]{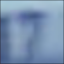} &
          \includegraphics[width=1.5cm]{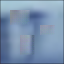} \\
            \includegraphics[width=1.5cm]{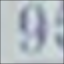} &
          \includegraphics[width=1.5cm]{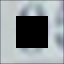} &
          \includegraphics[width=1.5cm]{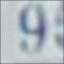} &
          \includegraphics[width=1.5cm]{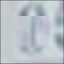} &
          \includegraphics[width=1.5cm]{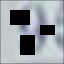} &
          \includegraphics[width=1.5cm]{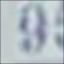} &
          \includegraphics[width=1.5cm]{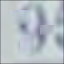} \\         
            \includegraphics[width=1.5cm]{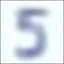} &
          \includegraphics[width=1.5cm]{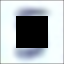} &
          \includegraphics[width=1.5cm]{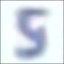} &
          \includegraphics[width=1.5cm]{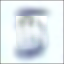} &
          \includegraphics[width=1.5cm]{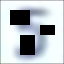} &
          \includegraphics[width=1.5cm]{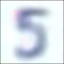} &
          \includegraphics[width=1.5cm]{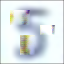} \\  
                      \includegraphics[width=1.5cm]{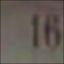} &
          \includegraphics[width=1.5cm]{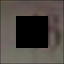} &
          \includegraphics[width=1.5cm]{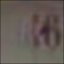} &
          \includegraphics[width=1.5cm]{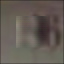} &
          \includegraphics[width=1.5cm]{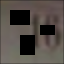} &
          \includegraphics[width=1.5cm]{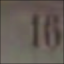} &
          \includegraphics[width=1.5cm]{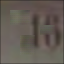} \\  

         Original & Masked & Ours & SIMDGM & Masked & Ours & SIMDGM
    \end{tabular}
\caption{Inpainting results on the SVHN dataset: Qualitative comparison with the method \cite{11} (fourth and seventh columns, referenced as SIMDGM), using the two masks shown in the second and fifth columns, is also displayed.}
\label{fig:FinalResultsSVHN}
\end{figure*}

\paragraph{Qualitative Assessment}
We separately analyze each step of our algorithm: The training of the generative model and the minimization procedure to infer the missing content. Since the inpainting optimum of the latter strongly depends on what the generative model is able to produce, a good estimation of the data latent space is crucial for our task. Figure \ref{fig:generatedDatasets} 
shows some images generated by our generative model trained with the CelebA and SVHN, 
respectively. Notice that the CelebA dataset is better estimated due to the fact that the number of images as well as the diversity of the dataset directly affects the prediction of the latent space and the estimated underlying probability density function (pdf). 
In contrast, as bigger the variability of the dataset, more spread is the pdf which difficult its estimation.

\begin{figure}
    \centering
    \begin{tabular}{cccc}
            \includegraphics[width=1.4cm]{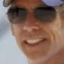} &
          \includegraphics[width=1.4cm]{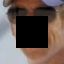} &
          \includegraphics[width=1.4cm]{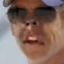} &
          \includegraphics[width=1.4cm]{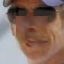} \\
            \includegraphics[width=1.4cm]{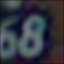} &
          \includegraphics[width=1.4cm]{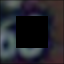} &
          \includegraphics[width=1.4cm]{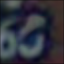} &
          \includegraphics[width=1.4cm]{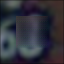} 
          \\
         Original & Masked & Ours & SIMDGM 
    \end{tabular}
\caption{Some examples of failure cases}
\label{fig:FailureCases}
\end{figure}

To evaluate our inpainting method we compare it with 
%
the semantic inpainting method of  \cite{11}. Some qualitative results are displayed in Figures \ref{fig:FinalResultsCelebA} and \ref{fig:FinalResultsSVHN}. 
Focusing on the CelebA results (Figure \ref{fig:FinalResultsCelebA}), obviously \cite{11} performs much better than local and non-local methods (Figure \ref{fig:f2}) since it also makes use of generative models. However, although that method is able to recover the semantic information of the image and infer the content of the missing areas, in some cases it keeps producing results with lack of structure and detail which can be caused either by the generative model or by the procedure to search the closest encoding in the latent space. We will further analyze it in the next section within the ablation study of our contributions. Since our method takes into account not only the pixel values but also the structure of the image this kind of problems are solved. In many cases, our results are as realistic as the real images. 
Notice that challenging examples, such as the fifth image from Figure \ref{fig:FinalResultsCelebA}, which image structures are not well defined, are not properly recovered with our method nor with \cite{11}. Some failure examples are shown in Figure \ref{fig:FailureCases}. 

Regarding the results on SVHN dataset (Figure \ref{fig:FinalResultsSVHN}), although they are not as realistic as the CelebA ones, the missing content is well recovered even when different numbers may semantically fit the context. As mentioned before, the lack of detail is probably caused by the training stage, due to the large variability of the dataset (and the small number of examples). Despite of this, let us notice that our qualitative results outperform the ones of \cite{11}. This may indicate that our algorithm is more robust in the case of smaller datasets than \cite{11}.

\paragraph{Quantitative Analysis and Evaluation Metrics}
The goal of semantic inpainting is to fill-in the missing information with realistic content. However, with this purpose, there are many correct possibilities to semantically fill the missing information. In other words, a reconstructed image equal to the ground truth would be only one of the several potential solutions. Thus, in order to quantify the quality of our method in comparison with other methods, we use different evaluation metrics: First, metrics based on a distance with respect to the ground truth and, second, a perceptual quality measure that is acknowledged to agree with similarity perception in the human visual system.
\begin{table*}[h]
\caption{
    Quantitative inpainting results for the central square mask (shown in Fig. \ref{fig:datasetMask}(c)-left), including an ablation study of our contributions in comparison with \cite{11}. The best results for each dataset are marked in bold and the best results for each method are underlined.}\label{table:central_bloc}
\centering 
    \resizebox{\textwidth}{!}{\begin{tabular}{| l | c c c | c c c | }
    \cline{2-7}
    \multicolumn{1}{c|}{} & \multicolumn{3}{c|}{CelebA dataset} & \multicolumn{3}{c|}{SVHN dataset}\\
    \hline
    Loss formulation &   MSE & PSNR & SSIM & MSE & PSNR & SSIM \\ 
    \hline
    \cite{11}  & 872.8672 & 18.7213 & 0.9071 & 1535.8693 & 16.2673  & 0.4925\\ 
    \cite{11} adding gradient loss  with $\alpha=0.1$, $\beta=0.9$ and $\eta=1.0$ & 832.9295 & 18.9247 & 0.9087 & 1566.8592 & 16.1805  & 0.4775 \\ 
    \cite{11} adding gradient loss  with $\alpha=0.5$, $\beta=0.5$ and $\eta=1.0$ & 862.9393 & 18.7710 & 0.9117 & 1635.2378 & 15.9950  & 0.4931 \\ 
          \cite{11} adding gradient loss  with $\alpha=0.1$, $\beta=0.9$ and $\eta=0.5$  & \underline{794.3374} & \underline{19.1308} & \underline{0.9130} & \underline{1472.6770} & \underline{16.4438}  & \underline{0.5041}  \\ 
    \cite{11} adding gradient loss  with $\alpha=0.5$, $\beta=0.5$ and $\eta=0.5$    & 876.9104 & 18.7013 & 0.9063 & 1587.2998 & 16.1242 & 0.4818 \\ 
    Our proposed loss with $\alpha=0.1$, $\beta=0.9$ and $\eta=1.0$ & 855.3476 & 18.8094 & 0.9158 & 631.0078 & 20.1305 & \underline{\textbf{0.8169}} \\ 
    Our proposed loss with $\alpha=0.5$, $\beta=0.5$ and $\eta=1.0$   & \underline{\textbf{785.2562}} & \underline{\textbf{19.1807}} & \underline{\textbf{0.9196}} & 743.8718 & 19.4158 & 0.8030\\ 
    Our proposed loss with $\alpha=0.1$, $\beta=0.9$ and $\eta=0.5$    & 862.4890 & 18.7733 & 0.9135 & \underline{\textbf{622.9391}} & \underline{\textbf{20.1863}} & 0.8005 \\ 
    Our proposed loss with $\alpha=0.5$, $\beta=0.5$ and $\eta=0.5$     & 833.9951 & 18.9192 & 0.9146 & 703.8026 & 19.6563  & 0.8000 \\ 
   \hline
    \end{tabular}}
\end{table*}

\begin{table*}[h]
\caption{
   Quantitative inpainting results for the three squares mask (shown in Fig. \ref{fig:datasetMask}(c)-right), including an ablation study of our contributions and a complete comparison with \cite{11}. The best results for each dataset are marked in bold and the best results for each method are underlined.}\label{table:treeSquares}
\centering 
    \resizebox{\textwidth}{!}{\begin{tabular}{| l | c c c | c c c | }
    \cline{2-7}
    \multicolumn{1}{c|}{} & \multicolumn{3}{c|}{CelebA dataset} & \multicolumn{3}{c|}{SVHN dataset}\\
    \hline
    Method &   MSE & PSNR & SSIM & MSE & PSNR & SSIM \\ 
    \hline
    \cite{11}  & 622.1092 & 20.1921 & 0.9087 & 1531.4601 &  16.2797 & 0.4791 \\ 
    \cite{11} adding gradient loss  with $\alpha=0.1$, $\beta=0.9$ and $\eta=1.0$ & 584.3051 & 20.4644 & 0.9067 & 1413.7107 & 16.6272 & 0.4875 \\ 
    \cite{11} adding gradient loss  with $\alpha=0.5$, $\beta=0.5$ and $\eta=1.0$ & 600.9579 & 20.3424 & 0.9080 & 1427.5251 & 16.5850  & 0.4889 \\ 
         \cite{11} adding gradient loss  with $\alpha=0.1$, $\beta=0.9$ and $\eta=0.5$    & 580.8126 & 20.4904 & 0.9115 & 1446.3560 & 16.5281  & \underline{0.5120} \\ 
    \cite{11} adding gradient loss  with $\alpha=0.5$, $\beta=0.5$ and $\eta=0.5$    & \underline{563.4620} & \underline{20.6222} & 0.9103 & \underline{1329.8546} & \underline{16.8928}  & 0.4974 \\ 
    Our proposed loss with $\alpha=0.1$, $\beta=0.9$ and $\eta=1.0$ & 424.7942 & 21.8490 & 0.9281 & 168.9121 & 25.8542 & 0.8960 \\ 
    Our proposed loss with $\alpha=0.5$, $\beta=0.5$ and $\eta=1.0$   & 380.4035 & 22.3284 & 0.9314 & 221.7906 & 24.6714 & \underline{\textbf{0.9018}} \\ 
    Our proposed loss with $\alpha=0.1$, $\beta=0.9$ and $\eta=0.5$    & \underline{\textbf{321.3023}} & \underline{\textbf{23.0617}} & \underline{\textbf{0.9341}} & \underline{\textbf{154.5582}} & \underline{\textbf{26.2399}}  & 0.8969 \\ 
    Our proposed loss with $\alpha=0.5$, $\beta=0.5$ and $\eta=0.5$     & 411.8664 & 21.9832 & 0.9292 & 171.7974 & 25.7806  & 0.8939  \\ 
   \hline
     \end{tabular}}
\end{table*}

In the first case, considering the real images from the database as the ground truth reference, the most used evaluation metrics are the Peak Signal-to-Noise Ratio (PSNR) and the Mean Square Error (MSE). 
Notice, that both MSE and PSNR, will choose as best results the ones with pixel values closer to the ground truth. 
In the second case, in order to evaluate perceived quality, we use the Structural Similarity index (SSIM) \cite{SSIM} 
used to measure the similarity between two images. It is considered to be correlated with the quality perception of the human visual system and is defined as:
\begin{equation}
\label{eq:SSIM}
\begin{split}
SSIM(x,y) = l(x,y) \cdot c(x,y) \cdot s(f,g) \\
\text{ where } \begin{cases}
               l(x,y) = \frac{2 \mu_x \mu_y  +C_1}{\mu_x^2 + \mu_g^2 + C_1}\\
               c(x,y) = \frac{2 \sigma_x \sigma_y  +C_2}{\sigma_x^2 + \sigma_g^2 + C_2}\\
               s(x,y) = \frac{2 \sigma_{xy} +C_3}{\sigma_x\sigma_y + C_3}
            \end{cases}
\end{split}
\end{equation}
The first term in (\ref{eq:SSIM}) is the luminance comparison function which measures the closeness of the two images mean luminance ($\mu_x$ and $\mu_y$). The second term is the contrast comparison function which measures the closeness of the contrast of the two images, where $\sigma_x,\sigma_y$ denote the standard deviations.  The third term is the structure comparison function which measures the correlation between $x$ and $y$. $C_1,C_2$ and $C_3$ are small positive constants avoiding dividing by zero. Finally, $\sigma_{xy}$ denotes the covariance between $x$ and $y$. The SSIM is maximal when is equal to one.

Given these metrics we compare our results with the one proposed by \cite{11} as it is the method more similar to ours. Tables \ref{table:central_bloc} and \ref{table:treeSquares} show the numerical performance of our method and \cite{11} using both the right and left inpainting masks shown in Figure \ref{fig:datasetMask}(c), respectively, named from now on, central square and three squares mask, respectively.
To perform an ablation study of all our contributions and a complete comparison with \cite{11}, Tables \ref{table:central_bloc} and \ref{table:treeSquares} not only show the results obtained by their original algorithm and our proposed algorithm, but also the results obtained by adding our new gradient-based term $\mathcal{L}_g(z|y,M)$ to their original inpainting loss. We present the results varying the trade-off effect between the different loss terms. 

Our algorithm always performs better than the semantic inpainting method by \cite{11}.  For the case of the CelebA dataset, the average MSE obtained by \cite{11} is equal to 872.8672 and 622.1092, respectively, compared to our results that are equal to 785.2562 and 321.3023, respectively. It is highly reflected in the results obtained using the SVHN dataset, where the original version of \cite{11} obtains an MSE equal to 1535.8693 and 1531.4601, using the central and three squares mask respectively, and our method 622.9391 and 154.5582. On the one side, the proposed WGAN structure is able to create a more realistic latent space and, on the other side, the proposed loss takes into account essential information in order to recover the missing areas. 

Regarding the accuracy results obtained with the SSIM measure, we can see that ours results always have a better perceived quality than the ones obtained by \cite{11}. In some cases, the values are close to the double, for example, in the case of using the dataset SVHN. 

In general, we can also conclude that our method is more stable in smaller datasets such in the case of SVHN. In our case, decreasing the number of samples in the dataset does not mean to reduce the quality of the inpainted images. Contrary with what is happening in the case of \cite{11}. Finally, in the cases where we add the proposed loss to the algorithm proposed by \cite{11}, in most of the cases the MSE, PSNR and SSIM improves. This fact clarifies the big importance of the gradient loss in order to perform semantic inpainting.

\section{\uppercase{Conclusions}} 
\label{sec:conclusions}
\noindent 
In this work we propose a new method that takes advantage of generative adversarial networks to perform  semantic inpainting in order to recover large missing areas of an image. 
This is possible thanks to, first, an improved version of the Wasserstein Generative Adversarial Network which is trained to learn the latent data manifold. Our proposal includes a new generator and discriminator architectures having stabilizing properties. Second, we propose a new optimization loss in the context of semantic inpainting which is able to properly infer the missing content by conditioning to the available data on the image, through both the pixel values and the image structure, while taking into account the perceptual realism of the complete image. Our qualitative and quantitative experiments demostrate that the proposed method can infer more meaningful content for incomplete images than local, non-local and semantic inpainting methods. In particular, 
our method qualitatively and quantitatively outperforms the related semantic inpainting method \cite{11} obtaining images with sharper edges, which looks like more natural and perceptually similar to the ground truth.

Unsupervised learning needs enough training data to learn the distribution of the data and generate realistic images to eventually succeed in semantic inpainting. A huge dabaset with higher resolution images would be needed to apply our method to more complex and diverse world scenes. The presented results are based on low resolution images (64x64 pixel size) and thus the inpainting method is limited to images of that resolution. Also, more complex features needed to represent such complex and diverse world scenes would require a deeper architecture.
Future work will follow these guidelines. 

\section*{\uppercase{Acknowledgements}}
The authors acknowledge partial support by MINECO/FEDER UE project, reference TIN2015-70410-C2-1 and by H2020-MSCA-RISE-2017 project, reference 777826 NoMADS.\\*
\bibliographystyle{apalike}
{\small
\bibliography{bib}}



\vfill
\end{document}